\documentclass[journal]{IEEEtran}
\ifCLASSINFOpdf
\else
\fi
\usepackage{amsmath}
\usepackage[multi-part-units=single]{siunitx}
\usepackage{graphicx}
\usepackage{multirow}
\usepackage{bm}
\usepackage{cite}

\begin{document}
%
\title{Automatic Plaque Detection in IVOCT Pullbacks Using Convolutional Neural Networks}
%
%
%

\author{Nils~Gessert$^1$,
		Matthias~Lutz$^1$,
        Markus~Heyder,
        Sarah~Latus,
        David~M.~Leistner,
        Youssef~S.~Abdelwahed,
        and~Alexander~Schlaefer
\thanks{$^1$ Both authors contributed equally.}        
\thanks{N. Gessert, M. Heyder, S. Latus and A. Schlaefer are with the Institute
of Medical Technology, Hamburg University of Technology,
Am-Schwarzenberg-Campus 3, 21073, Hamburg, Germany (e-mail: \{nils.gessert,markus.heyder,sarah.latus,schlaefer\}@tuhh.de).}
\thanks{M. Lutz is with the Universit\"atsklinikum Schleswig-Holstein, Arnold-Heller-Straße 3, 24105 Kiel, Germany (e-mail: matthias.lutz@uksh.de).}
\thanks{D. M. Leistner and Y. S. Abdelwahred are with the Charit\'{e} – Universit\"atsmedizin Berlin, Hindenburgdamm 30, 12203 Berlin, Germany (e-mail: \{david-manuel.leistner,youssef.abdelwahed\}@charite.de).}
\thanks{Copyright (c) 2017 IEEE. Personal use of this material is permitted. However, permission to use this material for any other purposes must be obtained from the IEEE by sending a request to pubs-permissions@ieee.org.}}

%
%

\markboth{IEEE TRANSACTIONS ON MEDICAL IMAGING}
{Gessert \MakeLowercase{\textit{et al.}}: Automatic Plaque Detection in IVOCT Images Using Convolutional Neural Networks}
%



\maketitle

\begin{abstract}

Coronary heart disease is a common cause of death despite being preventable. To treat the underlying plaque deposits in the arterial walls, intravascular optical coherence tomography can be used by experts to detect and characterize the lesions. In clinical routine, hundreds of images are acquired for each patient which requires automatic plaque detection for fast and accurate decision support. So far, automatic approaches rely on classic machine learning methods and deep learning solutions have rarely been studied. Given the success of deep learning methods with other imaging modalities, a thorough understanding of deep learning-based plaque detection for future clinical decision support systems is required. We address this issue with a new dataset consisting of in-vivo patient images labeled by three trained experts. Using this dataset, we employ state-of-the-art deep learning models that directly learn plaque classification from the images. For improved performance, we study different transfer learning approaches. Furthermore, we investigate the use of cartesian and polar image representations and employ data augmentation techniques tailored to each representation. We fuse both representations in a multi-path architecture for more effective feature exploitation. Last, we address the challenge of plaque differentiation in addition to detection. Overall, we find that our combined model performs best with an accuracy of $\SI{91.7}{\percent}$, a sensitivity of $\SI{90.9}{\percent}$ and a specificity of $\SI{92.4}{\percent}$. Our results indicate that building a deep learning-based clinical decision support system for plaque detection is feasible. 

\end{abstract}


%
\IEEEpeerreviewmaketitle

\section{Introduction}

Coronary heart disease is a common cause of death despite being preventable and treatable. Early detection of atherosclerotic plaque deposits can be critical for appropriate treatment. Intravascular optical coherence tomography (IVOCT) has established itself as an alternative to intravascular ultrasound (IVUS) \cite{jang2002visualization,tearney2012consensus} as an imaging modality for invasive coronary angiography. Compared to IVUS, IVOCT is capable of resolving microstructures in the arterial wall which is traded off against a smaller field of view (FOV) \cite{kini2017fibrous}. During the procedure, a catheter containing an OCT probe is inserted into a coronary artery. For acquisition, the probe is rotated and pulled back automatically, providing several hundred or thousands of high resolution cross-sectional images of the arterial walls which guide the practitioner's treatment decision process during an intervention. The assessment of these images requires extensive training of the interventionalist. Also, the vast amount of images acquired in a single pullback cannot all be reviewed by the practitioner during clinical routine. Therefore, various approaches for automatic IVOCT data analysis have been proposed. For example, automatic lumen segmentation \cite{tsantis2012automatic, roy2016lumen} and stent detection \cite{lu2012automatic,wang20153} have been proposed.  Also, automatic detection of plaque deposits has been proposed. The general feasibility of inferring plaque characteristics from IVOCT data has been addressed by correlating the images to histology data \cite{yabushita2002characterization}. IVOCT has been used to measure the thickness of fibrous caps in the arterial wall to assess the risk of ruptures and potential subsequent myocardial infarction \cite{wang2012volumetric,zahnd2015quantification}. Also, calcified plaque regions have been quantified as they can increase the risk of stenosis \cite{mehanna2013volumetric,mintz2015intravascular}. In general, the backscattering and attenuation coefficient of the OCT signal have been used frequently for identification of atherosclerotic tissue \cite{xu2008characterization,van2010atherosclerotic,vermeer2014depth,gargesha2015parameter,liu2017tissue}. Also, automatic classification methods using texture and optical properties as features have been proposed for plaque detection and segmentation \cite{ughi2013automated,celi2014vivo}. The segmentation of calcified plaque regions has been proposed using \textit{k}-means clustering and random forests \cite{athanasiou2014methodology}.
More recently, deep learning methods have been proposed for IVOCT data processing, e.g. for tissue layer segmentation \cite{abdolmanafi2017deep}. Deep learning models, in particular convolutional neural networks (CNNs), have shown remarkable success in image processing tasks such as classification \cite{Krizhevsky.2012} and segmentation \cite{Long.2015}. Also for medical tasks, such as disease classification \cite{esteva2017dermatologist} or tumor segmentation \cite{pereira2016brain}, deep learning methods have shown significant improvment over classical methods \cite{litjens2017survey}.
For IVOCT-based learning problems, CNNs have been employed, mostly in preliminary studies for plaque segmentation. Patch-wise approaches have been presented where a CNN predicts the class of the patch center pixel \cite{abdolmanafi2017deep,he2018convolutional}. Following the trend of dense segmentation architectures for improved performance \cite{Long.2015}, SegNet \cite{badrinarayanan2017segnet} has been employed for calcified plaque segmentation \cite{oliveira2018coronary}.

\begin{figure*}
\centering
\includegraphics[width=1\textwidth]{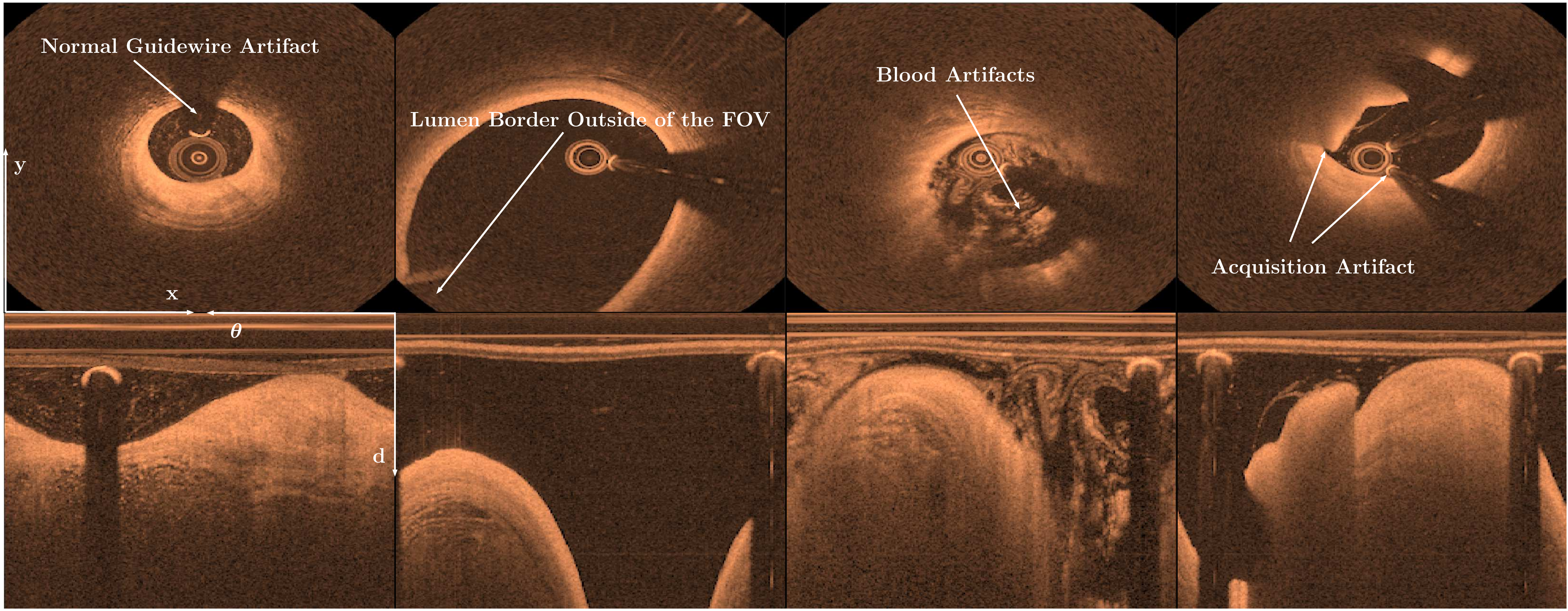}
\caption{Example IVOCT images, highlighting the difficult nature of the modality. The bottom row shows polar representations, the upper row shows the corresponding, transformed, cartesian images.}
\label{fig:example}
\end{figure*}

Although these approaches provide promising initial results, there are various shortcomings to be addressed. First of all, it is unclear whether learning from polar or cartesian representations is the better choice. Using cartesian images \cite{he2018convolutional} is more intuitive as their content resembles the anatomical structure of the artery. However, polar representations \cite{abdolmanafi2017deep,oliveira2018coronary} theoretically contain the same information while not being distorted by circular interpolation artifacts. For this reason, most classic approaches rely on polar images for image processing \cite{boi2018survey}. However, it is unclear whether either or both representations are more suitable for deep learning-based feature extraction.
Also, IVOCT acquisition is often accompanied by various artifacts, see Fig.~\ref{fig:example}. Many approaches rely on extensive preprocessing, including lumen segmentation and flattening \cite{rico2016automatic,roy2016lumen} which is hindered by artifacts that block light penetration of the arterial walls. As these artifcats are common in daily clinical use, automatic procedures should be sufficiently robust.
Moreover, previous approaches \cite{he2018convolutional,oliveira2018coronary,abdolmanafi2017deep} are limited in terms of dataset size and model choice (e.g. AlexNet \cite{Krizhevsky.2012}). More recent standard architectures such as ResNet \cite{He.2016} and Inception \cite{Szegedy.2016b} have been shown to significantly improve performance for medical learning tasks \cite{litjens2017survey}.
Also, transfer learning from ImageNet has been shown to be advantageous for medical learning problems \cite{shin2016deep}. For IVOCT-based tissue layer segmentation, rudimentary transfer learning with a partially frozen network has been shown \cite{abdolmanafi2017deep}. Another approach considered full fine-tuning for plaque classification \cite{gessert2018}. However, the optimal level of meaningful feature transfer and partial retraining has not been explored for IVOCT data. Previously, it has been shown that the optimal choice or transfer strategy varies for different modalities and problems \cite{tajbakhsh2016convolutional} which motivates a thorough investigation for IVOCT data.
Lastly, previous methods largely relied on segmentation which is a necessary step for treatment planning in many medical applications \cite{litjens2017survey}. However, for IVOCT interventions, the key problem is the large number of slices in every pullback. Thus, a clinical decision support system needs to provide feedback on the existence and characteristics of plaque deposits on the pullback level. Overall, a goal of such a system is to ensure that no deposit region remains undetected. 
For this purpose, segmentation is not immediately required.

To overcome these shortcomings, we provide a thorough investigation of IVOCT-based deep learning using a newly built dataset with slice-level labels. Three trained experts provide a class for each slice, defining the plaque type that is present in the image. In this way, we can quickly build a large dataset containing a variety of challenging cases, as indicated in Fig.~\ref{fig:example}. Using this dataset, we investigate important aspects to be considered for deep learning with IVOCT data.  
First, we study the use of polar and cartesian image representations for deep feature learning. In particular, we use data augmentation techniques, tailored to each representation. To improve model performance, we employ transfer learning from ImageNet using the established models ResNet \cite{He.2016} and DenseNet \cite{huang2017densely}. As natural images are vastly different from IVOCT data, we provide an in-depth analysis of transfer learning, showing performance for different transfer strategies.
 Additionally, we build a new two-path architecture out of pretrained parts from other models that leverages features from both image representations. Overall, we consider binary labels "plaque" and "no plaque", as well as a further differentiation of "calcified plaque" from "fibrous/lipid plaque". This choice is also clinically motivated as calcified plaque deposits often require different tools for treatment, such as rotablation.
This paper is structured as follows. First, we introduce our dataset and methods in Section~\ref{sec:methods}. Second, we provide results in Section~\ref{sec:results}. Third, we discuss our results in Section~\ref{sec:discussion} and last, we conclude the paper in Section~\ref{sec:conclusion}.


\section{Methods} \label{sec:methods}

\begin{figure*}[!ht]
\centering
\includegraphics[angle=90,width=1\textwidth]{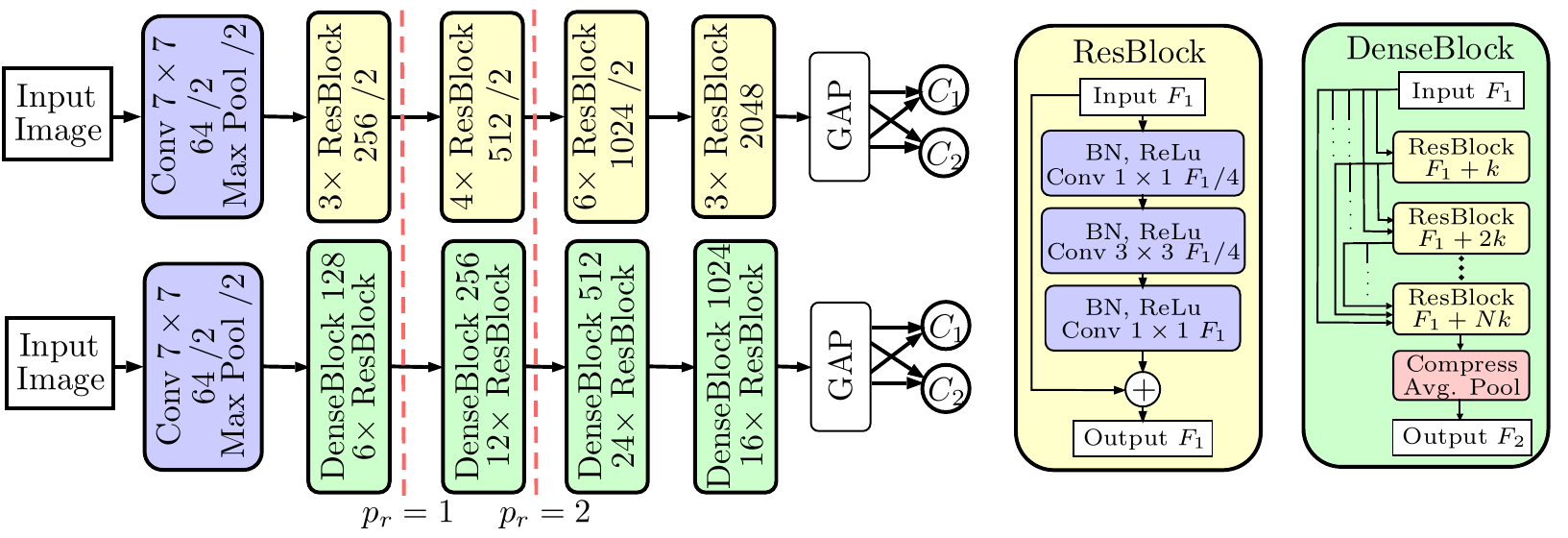}
\caption{The models we employ with our dataset. The upper model shows ResNet50-V2 \cite{He.2016b}, the lower model shows DenseNet121 \cite{huang2017densely}. In each ResBlock, input and output block, the number of output feature maps is given. $/2$ denotes spatial downsampling with a stride of two. The ResBlocks within the DenseBlock do not use the last $1\times 1$ convolution, instead, the previous convolution outputs feature maps of size $F_1$. Compress denotes a $1\times 1$ convolution that halves the number of feature maps. GAP denotes global average pooling \cite{Lin.2013}. Both architectures include dropout with a keep probability of $d=0.8$. $p_r$ denotes a retraining point for transfer learning. Left of the point, weights are freezed, right of the point, weights are retrained.}
\label{fig:models}
\end{figure*}

\subsection{Dataset} \label{sec:dataset}

We build a new dataset based on clinical IVOCT images that were acquired with a St. Jude Medical Ilumien OPTIS. The device is connected to a Dragonfly OPTIS imaging catheter which is inserted into the patient's coronary vessels. Inside the vessel, images are acquired by rotating the OCT probe inside the catheter and pulling it backwards. In this way, a continuous M-Scan consisting of 1D depth scans (A-Scans) is created. The A-Scans from each $\SI{360}{\degree}$ turn are arranged in a B-Scan, which is the polar representation shown in Fig.~\ref{fig:example}. The polar image $I_p(d,\theta)$ can be transformed into cartesian space with the transformation $x = d\cos(\theta)$ and $y = d\sin(\theta)$. Applying interpolation in between the transformed pixels results in the cartesian representation $I_c(x,y)$, which resembles a cross-sectional view of the artery, see Fig.~\ref{fig:example}. 

For ground-truth annotation, three trained experts with daily experience in IVOCT-assisted interventions determine the type of plaque that is present in a B-Scan. The experts provide labels on the B-Scan level, i.e. they assign the labels "no plaque", "calcified plaque" or "lipid/fibrous plaque" to each B-Scan presented to them. All experts were partially provided the same images and different images for labeling. For the same images, the final label is determined based on consensus between the experts, i.e. different opinions on the plaque type were resolved by asking the experts for a repetition of their evaluation. Providing different images to experts allows for a larger but potentially noisily labeled dataset. We sample our test set from the consensus set to ensure a meaningful evaluation. Before resolving conflicting evaluations we found an agreement of $\SI{87}{\percent}$ for binary differentiation and $\SI{68}{\percent}$ for multi-class differentiation. 

In total, the dataset consists of 4000 images from 49 patients. We split off an independent test set of 742 images from 9 patients. We report our results based on this test set. We perform three-fold cross-validation on the training set in order to select relevant hyperparameters that are introduced in Section~\ref{sec:training}. We use the ground-truth labels in two variants. For general results on plaque detection, we use binary labels, i.e. "plaque" and "no plaque". We refer to this dataset as the binary dataset. Furthermore, we consider a multi-type plaque dataset where we divide the class "plaque" further into "calcified plaque" and "lipid/fibrous plaque". This dataset serves the purpose of showing the feasibility of further plaque differentiation with clinical relevance.

For the binary dataset, plaque and non-plaque images are approximately balanced. For the mutli-type dataset, roughly \SI{10}{\percent} are labeled as "calcified plaque", \SI{40}{\percent} are labeled as "fibrous/lipid plaque" and \SI{50}{\percent} are labeled as "no plaque". None of the images contains a stent as this would distort results. Our goal is to detect untreated plaque deposits and a deep learning model might learn to detect stents instead of the underlying disease. 


\subsection{Preprocessing and Data Augmentation} \label{sec:data_aug}

The original polar images obtained from the OCT device have a resolution of $496 \times 960$ pixels. Thus, the transformed cartesian images have a resolution of $1920 \times 1920$ pixels. For training, we resize both to a resolution of $300 \times 300$, which matches the range used with standard architectures in the natural image domain \cite{Krizhevsky.2012,He.2016,huang2017densely}. Note, that this leads to the cartesian images having half the depth resolution of the polar images. 

For data augmentation, an intuitive choice is to use random rotations for the cartesian images due to their circular strucutre. We apply random rotations with $\alpha \in [0\si{\degree},360\si{\degree}]$ to cartesian images. Moreover, we apply random flipping along the $x$ and $y$ direction. For polar images, rotations are not meaningful. Instead, we shift the polar images randomly along the $\theta$ direction with $s_{\theta} \in [0 px, 300 px]$. If A-Scans are shifted out of the image along the positive $\theta$ direction, they are added back on the other side. In this way, we achieve a transformation that matches a rotation in cartesian space. Additionally, we apply random flipping along the $\theta$ direction. For both image representations, we apply random cropping to an image size of $270 \times 270$. For evaluation, we do not use any rotations or flipping and we use a center crop of size $270 \times 270$. 

\begin{figure*}[!ht]
\centering
\includegraphics[width=1\textwidth]{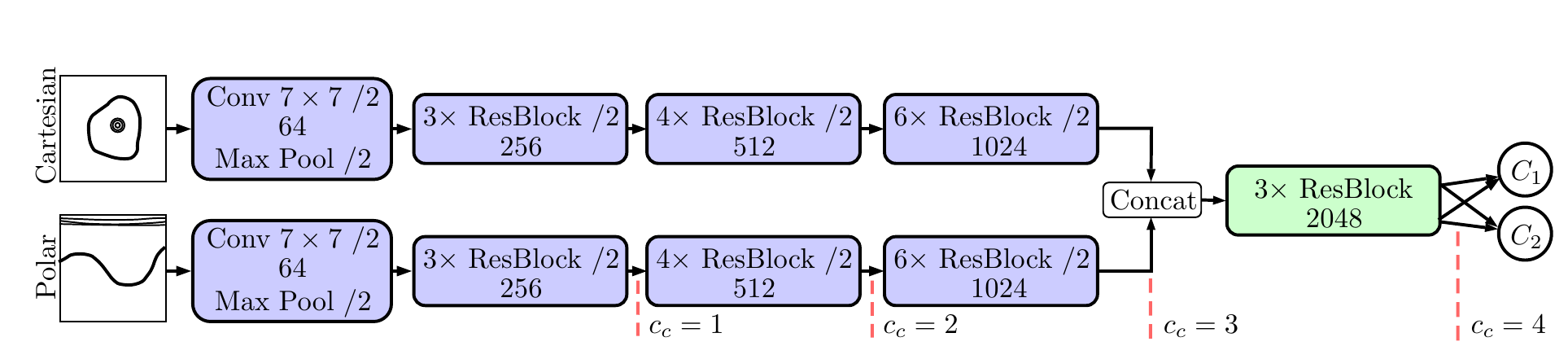}
\caption{The two-path architecture for simultaneous use of polar and cartesian representations, shown for the ResNet model. We initialize all weights with the pretrained values from ImageNet training. The point $c_c = i$ denotes feature concatenation after the $i^{th}$ ResNet block. Here, $c_c=3$ is shown.}
\label{fig:twopath}
\end{figure*}

\subsection{Model Architectures} \label{sec:models}

We employ two state-of-the-art architectures from the natural image domain, namely ResNet50-V2 \cite{He.2016} and DenseNet-121 \cite{huang2017densely}. The architectures' concepts are shown in Fig.~\ref{fig:models}. 

The ResNet model makes use of residual blocks for a better gradient flow and improved optimization. The key idea of this method is to learn a residual $\mathcal{F}(x) = \mathcal{H}(x)-x$ instead of the target mapping $\mathcal{H}(x)$. The resulting skip connections have been shown to improve gradient propagation through the network \cite{He.2016b}. Furthermore, bottlenecks are used which significantly reduce the number of parameters and the computational effort in the network. Here, a $1 \times 1$ convolution first downsamples the input tensor along the feature map dimension. Then, the normal $3 \times 3$ filter is applied on the lower dimensional embedding. Afterwards, a $1 \times 1$ filter upsamples the tensor once again to the original feature map size. An efficient architecture is particularly important for the problem at hand, as our dataset is relatively small which increases the risk of overfitting. 

The DenseNet model is focused on efficiency even more, as it relies on heavy feature reuse in each block. The idea of this architecture is to use all features from previous layers for current layer $l$. I.e., the output of the $l^{th}$ layer is computed as $x_l = H([x_0,x_1,...,x_{l-1}])$. In this way, the output feature maps of all previous layers are reused. This allows for overall significantly smaller feature maps and a reduced number of parameters. Due to this structure, the number of feature maps grows linearily with a growth rate $k$. Here, we use $k=32$. Furthermore, the architecture uses compression layers in between dense blocks in order to keep the feature map sizes low. A $1 \times 1$ convolution downsamples the feature maps by a factor $c$. Here, $c=0.5$ is used. This architecture also makes use of bottlenecks, as described for the ResNet model.

As our dataset is small compared to, e.g., ImageNet \cite{Krizhevsky.2012}, we make use of transfer learning which has been used successfully in the medical image domain \cite{shin2016deep}. 
Therefore, we consider different levels of weight freezing, both for the ResNet and DenseNet model. The retraining points are indicated in Fig.~\ref{fig:models}. E.g., for $p_r = 1$, we freeze all weights on the left of $p_r = 1$ and retrain the rest. Thus, we consider two partial retraining scenarios, training from scratch and full retraining for both models. In all cases, we remove the last fully-connected layer and replace it with a layer that has a matching number of outputs for our binary or multi-type labels. 

Furthermore, we consider a fusion of polar and cartesian images in a single architecture in order to investigate whether joint features from the two representations provide an overall benefit for the problem. Related to this approach, CNNs with multiple paths have been used for multi-resolution inputs \cite{Kamnitsas.2017} and differently deformed images \cite{gessert2018force}. So far, different image representations have not been combined in this way and there are no related approaches for IVOCT. The proposed architecture based on ResNet is shown in Fig.~\ref{fig:twopath}. Similarly, we construct the architecture for DenseNet. Each path receives the polar or cartesian representation as its input. We apply the same data augmentation techniques to each representation, as described in Section~\ref{sec:data_aug}. The point $c_c$ describes the concatenation point within the network, where the image features are fused. The concatenation is performed along the tensor's feature map dimension, i.e. the features from each path are stacked together. Assume a tensor size of $[B,H,W,F]$  before concatenation where $B$ is the batch size, $H$ and $W$ are the spatial height and width and $F$ is the number of feature maps. The concatenated feature maps have the shape $[B,H,W,2F]$. For ResNet, the concatenation is performed in between two ResBlocks. Note, that we use the following $1 \times 1$ reduction operation in the next ResBlock in order to reduce the feature map size back to the normal size that would occur in the single-path ResNet. For DenseNet, the concatenation is performed inside a DenseBlock, before the compression operation. Similar to ResNet, the compression block keeps the overall feature map size at a reasonable level despite the concatenation.
This choice also enables us to directly use the pretrained weights for most parts of the architecture as most of the weights have the same shape. We initialize both paths before concatenation with the normal pretrained weights. After the concatenation, we can also reuse the weights due to our previously described downsampling strategy. Only the compression operation itself comes with a different weight shape, that cannot be adapted immediately. Consider two feature maps before concatenation from the two path, each with shape $[B,H,W,F_1]$ where $B$ is the batch size, $H$ and $W$ are the spatial height and width and $F_1$ is the number of feature maps. Thus, the downsampling convolution's weight tensor has the shape $[1,1,2F_1,F_2]$ where $F_2$ is the downsampled feature map size. We assign the original, pretrained weight tensor of shape $[1,1,F_1,F_2]$ both to the sliced tensor of shape $[1,1,1...F_1,F_2]$ and $[1,1,F_1...2F_1,F2]$. We show that this initialization does have a significant impact on performance.

\subsection{Training} \label{sec:training}

For model training we minimize the cross-entropy loss which is typically used for classification tasks, i.e.

\begin{equation}
	\mathcal{L} = - \sum_{c=1}^{M} y_c \log{\hat{y}_c}
\end{equation}

where $y$ is the ground-truth label, $\hat{y}$ is the softmax-normalized model prediction and $M$ the number of classes. For the multi-class cases, we weight the loss for "calcified plaque" higher in order to counter class imbalance. For training, we use the Adam algorithm \cite{Kingma.2014} with a starting learning rate of $l_r = \num{e-4}$. To find the optimal schedule, we reduce the learning rate by a factor of two when the validation error saturates. We use a batch size of $B=30$ for single-path models and $B=20$ for two-path models. In total, we train each model for $300$ epochs. We find relevant hyperparameters with a restricted grid search with a coarse grid using the $F_1$-score from threefold cross-validation. These hyperparameters include the initial learning rate, dropout rate, input image size and crop size. Relevant results are reported for the independent test set.
We implement our models using Tensorflow \cite{Abadi.2016}.

\subsection{Experiments Overview} \label{sec:experiments}

First, we present results for binary classification, both for the Resnet and Densenet model, as well as for polar and cartesian images. We show results with and without our data augmentation strategy. All models are pretrained on ImageNet. 

Second, we investigate transfer learning by showing results without pretraining, and two partial retraining scenarios, both for Densenet and Resnet using cartesian images.

Third, we provide results for our two-path architecture, using both image representations. Results for Resnet and Densenet are presented. We show results for different concatenation points, investigating the optimal abstraction level for feature fusion. We present results for our weight initialization strategy at the concatenation point.

Last, we consider multi-class classification for both CNN models. Again, we consider both image representations and the use of data augmentation. All models are pretrained on ImageNet.

For evaluation of binary classification models, we report accuracy (Acc.), sensitivity (Sens.), specificity (Spec.), and the F1-score. For multi-class classification, we report the per-class weighted accuracy for each class and the F1-score with all classes. 
The per-class weighted accuracy is calculated by dividing the number of true positive examples by the number of all examples for that particular class.

\section{Results} \label{sec:results}

\begin{table}[!t]
\renewcommand{\arraystretch}{1.3}
\caption{Results for binary plaque classification. All models are pretrained on ImageNet and all weights are fine-tuned.}
\label{tab:bin_pre}
\centering
\begin{tabular}{l l l l l l}
 & & Acc. & Sens. & Spec. & F1-Score \\
\hline
\parbox[t]{2mm}{\multirow{4}{*}{\rotatebox[origin=c]{90}{Data Aug.}}} &  Densenet Cart. & $0.892$ & $\bm{0.874}$ & $0.907$ & $0.885$ \\
& Densenet Polar. & $0.871$ & $0.852$ & $0.883$ & $0.865$ \\
& \textbf{Resnet Cart.} & $\bm{0.903}$ & $0.861$ & $\bm{0.937}$ & $\bm{0.888}$ \\
& Resnet Polar & $0.872$ & $0.888$ & $0.859$ & $0.861$ \\
\hline
\parbox[t]{2mm}{\multirow{4}{*}{\rotatebox[origin=c]{90}{No Data Aug.}}} &  Densenet Cart. & $0.755$ & $0.693$ & $0.807$ & $0.714$ \\
& Densenet Polar. & $0.821$ & $0.802$ & $0.831$ & $0.818$ \\
& Resnet Cart. & $0.740$ & $0.776$ & $0.719$ & $0.737$ \\
& Resnet Polar & $0.814$ & $0.810$ & $0.854$ & $0.813$ \\
\hline
\end{tabular}
\end{table}

\begin{figure*}[!ht]
\centering
\includegraphics[width=1\textwidth]{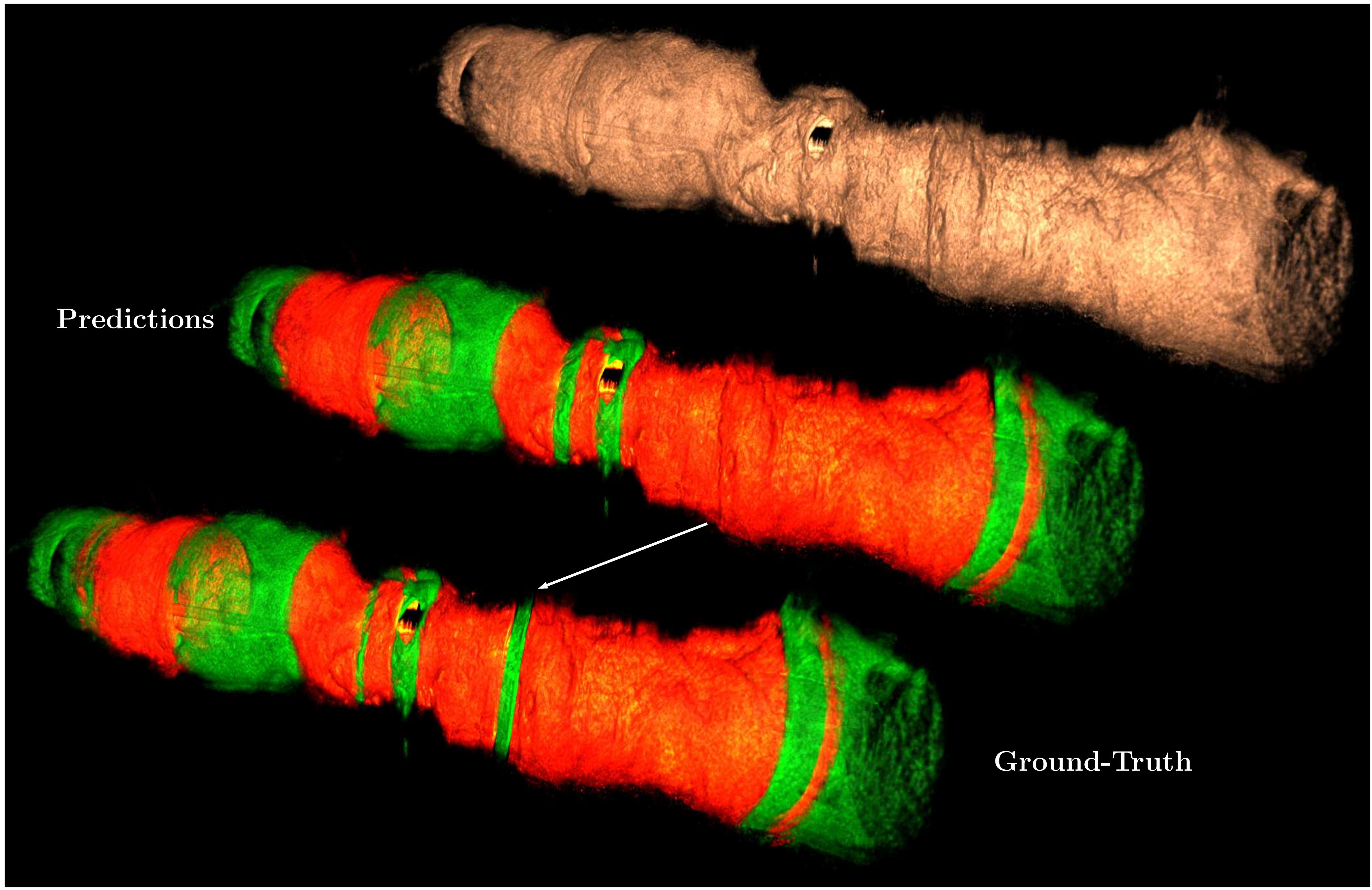}
\caption{Qualitative results shown with a rendered pullback. Top, the original pullback is shown. Mid, predicted plaque regions are shown along the pullback. Bottom, the ground-truth labels are shown along the pullback. The color red indicates regions with plaque and the color green indicates regions without plaque. The arrow indicates visibly incorrect predictions. Rendering was performed by stacking cartesian slices along the catheter's center.}
\label{fig:vis_pullback}
\end{figure*}

For binary plaque classification, results are shown in Table~\ref{tab:bin_pre}. The most evident difference is visible for data augmentation strategies. Also, the performance improvement through data augmentation is notably higher for cartesian images. Furthermore, we can observe that cartesian images lead to a better performance than polar images. The difference between the Resnet and Densenet model is comparatively small. Additionally, we show qualitative results for plaque detection along a pullback. The visualization is shown in Fig.~\ref{fig:vis_pullback}. Although some slices are missclassified, the system overall provides a good overview on the presence of plaque along the pullback. 

Furthermore, we investigate different transfer learning strategies. Table~\ref{tab:trans_res} shows the results for different levels of retraining. We consider training from scratch, full fine-tuning and two different partial freezing scenarios. Overall, fine-tuning all weights within the network performs best. This is followed by partially freezing some weights. The more weights are frozen, the more performance deteriorates. Performance is worst, when no pretraining is performed at all. We also report results for training from scratch with both cartesian and polar images. The difference between full retraining and training from scratch is larger for cartesian images.

\begin{table}[!t]
\renewcommand{\arraystretch}{1.3}
\caption{Results for different transfer learning scenarios using cartesian images, unless indicated otherwise. Full refers to full fine-tuning of all weights. From scratch refers to training with random weight initialization instead of pretraining. $p_r$ denotes the point of weight freezing in the network, see Fig.~\ref{fig:models}.}
\label{tab:trans_res}
\centering
\begin{tabular}{l l l l l l}
 & & Acc. & Sens. & Spec. & F1-Score \\
\hline
\parbox[t]{2mm}{\multirow{4}{*}{\rotatebox[origin=c]{90}{Densenet}}} & Full & $0.892$ & $\bm{0.874}$ & $0.907$ & $0.885$ \\
& From Scratch & $0.737$ & $0.761$ & $0.718$ & $0.721$ \\
& $p_r=1$ & $0.861$ & $0.842$ & $0.891$ & $0.866$ \\
& $p_r=2$ & $0.848$ & $0.817$ & $0.892$ & $0.840$ \\
& From Scratch (Polar) & $0.758$ & $0.770$ & $0.743$ & $0.751$ \\
\hline
\parbox[t]{2mm}{\multirow{4}{*}{\rotatebox[origin=c]{90}{Resnet}}} & \textbf{Full} & $\bm{0.903}$ & $0.861$ & $\bm{0.937}$ & $\bm{0.888}$ \\
& From Scratch & $0.758$ & $0.733$ & $0.770$ & $0.716$ \\
& $p_r=1$ & $0.882$ & $0.870$ & $0.893$ & $0.868$ \\
& $p_r=2$ & $0.851$ & $ 0.785$ & $0.894$ & $0.823$ \\
& From Scratch (Polar) & $0.774$ & $0.765$ & $0.800$ & $0.757$ \\
\hline
\end{tabular}
\end{table}

Moreover, we consider the two-path architecture introduced in Section~\ref{sec:models}. The results for the approach are shown in Table~\ref{tab:two_path_res}. We show results for different concatenation points and for our weight initialization strategy at the concatenation point. The model with an intermediate concatenation point $c_c = 3$ performs best, being close to $c_c = 4$. Moving the point closer to the input of the network deteriorates performance. Moreover, the proposed weight initialization strategy considerably improves performance.

\begin{table}[!t]
\renewcommand{\arraystretch}{1.3}
\caption{Results for the two-path architecture concept, both for Densenet and Resnet. $c_c$ denotes the concatenation point in the network, see Fig.~\ref{fig:twopath}. No Init. refers to no use of our weight initialization strategy. All other weights are still initialized with their pretrained values. We demonstrate it with the best performing model with concatenation point $c_c=3$.}
\label{tab:two_path_res}
\centering
\begin{tabular}{l l l l l l}
 & & Acc. & Sens. & Spec. & F1-Score \\
\hline
\parbox[t]{2mm}{\multirow{4}{*}{\rotatebox[origin=c]{90}{Densenet}}} & No Init. & $0.853$ & $0.836$ & $0.875$ & $0.846$ \\
& $c_c=2$ & $0.871$ & $0.843$ & $0.895$ & $0.864$ \\
& $c_c=3$ & $0.910$ & $0.892$ & $0.919$ & $0.908$ \\
& $c_c=4$ & $0.903$ & $0.881$ & $0.923$ & $0.895$ \\
\hline
\parbox[t]{2mm}{\multirow{4}{*}{\rotatebox[origin=c]{90}{Resnet}}} & No Init. & $0.871$ & $0.850$ & $0.883$ & $0.862$ \\
& $c_c=2$ & $0.867$ & $0.871$ & $0.852$ & $0.856$ \\
& $\bm{c_c=3}$ & $\bm{0.917}$ & $\bm{0.909}$ & $\bm{0.924}$ & $\bm{0.913}$ \\
& $c_c=4$ & $0.904$ & $0.899$ & $0.906$ & $0.901$ \\
\hline
\end{tabular}
\end{table}

Last, we show multi-class plaque classification results in Table~\ref{tab:multi_pre}. Overall, the classification is worse than for the binary classification case. Still, classification of calcified plaque achieves a slightly lower but similar performance as for the detection of other plaque types. The general trends in terms of performance are similar to binary classification. The use of cartesian images improves performance slightly which shows consistently for both models. No use of data augmentation significantly reduces classification performance.

\begin{table}[!t]
\renewcommand{\arraystretch}{1.3}
\caption{Results for multi-class plaque classification. The classes are "calcified plaque" ($c_1$), "fibrous/lipid plaque" ($c_2$) and "no plaque" ($c_3$). All models are pretrained on ImageNet and all weights are retrained. W.A. refers to the weighted per-class accuracy.}
\label{tab:multi_pre}
\centering
\begin{tabular}{l l l l l l}
 & & W.A. $c_1$ & W.A. $c_2$ & W.A. $c_3$. & F1-Score \\
\hline
\parbox[t]{2mm}{\multirow{4}{*}{\rotatebox[origin=c]{90}{Data Aug.}}} &  \textbf{Densenet Cart.} & $0.780$ & $\bm{0.848}$ & $\bm{0.897}$ & $\bm{0.833}$ \\
& Densenet Polar. & $0.755$ & $0.803$ & $0.867$ & $0.794$ \\
& Resnet Cart. & $\bm{0.794}$ & $0.822$ & $0.873$ & $0.829$ \\
& Resnet Polar & $0.762$ & $0.799$ & $0.856$ & $0.805$ \\
\hline
\parbox[t]{2mm}{\multirow{4}{*}{\rotatebox[origin=c]{90}{No Data Aug.}}} &  Densenet Cart. & $0.646$ & $0.702$ & $0.755$ & $0.708$ \\
& Densenet Polar. & $0.691$ & $0.737$ & $0.801$ & $0.757$ \\
& Resnet Cart. & $0.630$ & $0.689$ & $0.753$ & $0.694$ \\
& Resnet Polar & $0.687$ & $0.691$ & $0.770$ & $0.733$ \\
\hline
\end{tabular}
\end{table}

\section{Discussion} \label{sec:discussion}

With the recent success of deep learning methods for medical image analysis, future IVOCT-based clinical decision support systems require a profound understanding of deep learning with IVOCT data. Therefore, we present an in-depth investigation of deep learning-based plaque detection in IVOCT pullbacks. We built a new dataset that is specifically targeted at this task by using slice-level labels. 
This approach provides an assessment at the pullback level, which is particularly required for a potential clinical decision support system, see Fig.~\ref{fig:vis_pullback}. 
This approach allows us to quickly build a large dataset which enables us to consider a variety of difficult IVOCT images containing various artifacts as shown in Fig.~\ref{fig:example}, which is particularly relevant for clinical application. A robust detection model would avoid the necessity of repeated pullbacks in case of, e.g., minor acquisition or blood artifacts. Overall, Fig.~\ref{fig:vis_pullback} shows that our dataset can be used for plaque detection along pullbacks, as most regions are classified correctly. 

\textbf{Polar and cartesian images.} Using our dataset, we first investigate the use of polar and cartesian image representations for deep feature learning. For binary classification, we find that the best cartesian model achieves an accuracy of $0.903$ compared to $0.872$ for the best polar model. Overall, models using cartesian images appear to perform better, see Table~\ref{tab:bin_pre}. This is a surprising results as polar images should be richer in tissue property information since we downsampled both images to the same size of $300\times300$. Due to the circular transformation, the depth scans in the cartesian images only possess half the resolution compared to polar images. This indicates, that the relevant features for plaque detection are easier to exploit from cartesian images for deep learning models. It should be noted, that practitioners also generally use cartesian images for assessment. 

\textbf{Data augmentation.} We used extensive data augmentation, both for cartesian and polar images. In particular, rotations for cartesian images and shifting for polar images are well suited for IVOCT data. Considering the results in Table~\ref{tab:bin_pre}, cartesian-based models benefit substantially more from data augmentation. Accuracy improves by $\SI{16}{\percent}$ for cartesian images and $\SI{6}{\percent}$ for polar images. This shows that the superior performance of cartesian images is also largely tied to data augmentation. Our dataset is still small compared to standard sets being used in the natural image domain \cite{Krizhevsky.2012} and thus dependence on data augmentation can be expected. We highlight that it is a key aspect to consider when building a deep learning-based decision support system as performance heavily depends on it.

\textbf{Transfer learning.} We provide an extensive evaluation of transfer learning for IVOCT-based deep learning. The results in Table~\ref{tab:trans_res} show that there is a significant difference in performance between training from scratch and fine tuning. For both models, a performance gain of approximately $\SI{15}{\percent}$ was achieved. Previous transfer learning approaches for other medical imaging modalities achieved similar performance improvements \cite{shin2016deep}. For IVOCT, it is still notable that pretraining on natural images works well, considering that the modalities are vastly different. For further investigation, we considered partially freezing early layers of the architectures which is motivated by the fact that earlier layers tend to learn more generic features\cite{Yosinski.2014}. While our results show that performance is still high compared to training from scratch, accuracy deteriorates the more weights are frozen during training. This indicates that IVOCT features are very different from natural image features and thus, full retraining should be performed. This extends previous findings on transfer learning and IVOCT where only partially frozen networks were considered \cite{abdolmanafi2017deep}. Moreover, we compared how transfer learning affects performance for polar and cartesian representations. Pretraining appears to be more effective with cartesian images which might be related to the fact that the source domain images are also cartesian. Despite the significant difference between natural and IVOCT images, the shared cartesian structure appears to be beneficial for transfer learning.

\textbf{Multi-path architecture.} Besides representation-specific data augmentation we also considered a fusion of polar and cartesian image features for additional insights and performance gain. Overall, the two-path architectures with late feature concatenation slightly increase performance over the best single-path models by $\SI{1.4}{\percent}$ for Resnet and $\SI{1.8}{\percent}$ for Densenet, see Table~\ref{tab:bin_pre} and Table~\ref{tab:two_path_res}. First, this underlines that deep learning models are able to extract different features from polar and cartesian representations, as the performance increases when using both. Also, the learned features appear to differ earlier in the network, as early feature concatenation decreases performance. Second, we show an effective way of reusing pretrained weights for a modified architecture. As discussed before, transfer learning significantly boosts performance and thus pretraining is mandatory for high performance models. Our simple yet effective initialization strategy at the concatenation point shows that task-specific architecture design is still feasible when being tied to pretrained models. It should be noted that this concept works consistently across both the Resnet and Densenet model, although their structure and feature propagation mechanics are different.

\textbf{Multi-class classification.} Besides binary plaque detection, we consider a subdivision into calcified and fibrous/lipid plaque types. While fibrous and lipid plaques are relevant for risk assessment in terms of potential ruptures and thrombosis \cite{wang2012volumetric}, detection of calcified regions is relevant in terms of tool selection for treatment. Often, the use of rotablation or a cutting balloon is required for calcified regions. 
Our results in Table~\ref{tab:multi_pre} show that calcified plaques can be distinguished from others. 
However, the overall F1-scores are significantly lower than for binary classification despite both approaches being based on the same image data. This indicates uncertainty about specific plaque types. Considering our labeling strategy, this indicates that plaque differentiation is subjective and dependent on the labeling expert. This is not surprising in the sense that plaque development over time is continuous. E.g., the decision whether a plaque deposit is still to be considered fibrous or already calcified is not always clear and might change even within the same plaque region. Our results highlight this well-known issue \cite{tearney2012consensus} and suggest extended research towards both experts' labeling and multi-class learning approaches. The effect of continuous plaque changes along pullbacks is also visible in Fig.~\ref{fig:vis_pullback} for the binary case. The expert annotations indicate that a small area within a large region is free of plaque. As plaque develops continuous, the differences between slices are likely small. The experts made a fine-grained distinction which the model could not detect in this case which indicates that the expert's chosen threshold might have differed between training cases and this particular test case.

For all parts of our investigation we considered two state-of-the-art deep learning models ResNet and Densenet which showed consistent results for all experiments which underlines the validity of the insights gained from our experiments. 


\section{Conclusion} \label{sec:conclusion}

We present an in-depth analysis of plaque detection in IVOCT pullbacks using convolutional neural networks. For this purpose, we built a new database with IVOCT slices labeled by three trained experts. 
We employ two state-of-the-art  deep learning models, Resnet and Densenet, for binary and multi-class plaque classification from cartesian and polar IVOCT images. We show consistently across models, that cartesian images appear to be advantageous for deep learning. We develop data augmentation strategies, tailored to each image representation which result in a considerable performance boost. Additionally, we investigate transfer learning which provides high performance with model fine-tuning. Furthermore, we develop a multi-path model that fuses features from both representations for improved accuracy and effectively reuses weights from different, pretrained architectures. Last, we consider differentiation of calcified plaque from other plaque types which works well but also highlights potential issues of plaque differentiation and expert judgement. Overall, our results show that a deep learning-based clinical decision support system providing plaque detection along pullbacks would be feasible. 
For future work, the dataset can be extended in size as well as the number of experts consulted for assessment. Based on this, additional analysis with respect to plaque differentiation could be performed. Additionally, a decision support system could be integrated into catheter laboratories for the assessment of usefulness in clinical practice.



\ifCLASSOPTIONcaptionsoff
  \newpage
\fi



%


\bibliographystyle{IEEEtran} 
\bibliography{egbib}

\begin{thebibliography}{10}
\providecommand{\url}[1]{#1}
\csname url@samestyle\endcsname
\providecommand{\newblock}{\relax}
\providecommand{\bibinfo}[2]{#2}
\providecommand{\BIBentrySTDinterwordspacing}{\spaceskip=0pt\relax}
\providecommand{\BIBentryALTinterwordstretchfactor}{4}
\providecommand{\BIBentryALTinterwordspacing}{\spaceskip=\fontdimen2\font plus
\BIBentryALTinterwordstretchfactor\fontdimen3\font minus
  \fontdimen4\font\relax}
\providecommand{\BIBforeignlanguage}[2]{{%
\expandafter\ifx\csname l@#1\endcsname\relax
\typeout{** WARNING: IEEEtran.bst: No hyphenation pattern has been}%
\typeout{** loaded for the language `#1'. Using the pattern for}%
\typeout{** the default language instead.}%
\else
\language=\csname l@#1\endcsname
\fi
#2}}
\providecommand{\BIBdecl}{\relax}
\BIBdecl

\bibitem{jang2002visualization}
I.-K. Jang, B.~E. Bouma, D.-H. Kang, S.-J. Park, S.-W. Park, K.-B. Seung, K.-B.
  Choi, M.~Shishkov, K.~Schlendorf, E.~Pomerantsev \emph{et~al.},
  ``Visualization of coronary atherosclerotic plaques in patients using optical
  coherence tomography: comparison with intravascular ultrasound,''
  \emph{Journal of the American College of Cardiology}, vol.~39, no.~4, pp.
  604--609, 2002.

\bibitem{tearney2012consensus}
G.~J. Tearney, E.~Regar, T.~Akasaka, T.~Adriaenssens, P.~Barlis, H.~G. Bezerra,
  B.~Bouma, N.~Bruining, J.-m. Cho, S.~Chowdhary \emph{et~al.}, ``Consensus
  standards for acquisition, measurement, and reporting of intravascular
  optical coherence tomography studies,'' \emph{Journal of the American College
  of Cardiology}, vol.~59, no.~12, pp. 1058--1072, 2012.

\bibitem{kini2017fibrous}
A.~S. Kini, Y.~Vengrenyuk, T.~Yoshimura, M.~Matsumura, J.~Pena, U.~Baber,
  P.~Moreno, R.~Mehran, A.~Maehara, S.~Sharma \emph{et~al.}, ``Fibrous cap
  thickness by optical coherence tomography in vivo,'' \emph{Journal of the
  American College of Cardiology}, vol.~69, no.~6, pp. 644--657, 2017.

\bibitem{tsantis2012automatic}
S.~Tsantis, G.~C. Kagadis, K.~Katsanos, D.~Karnabatidis, G.~Bourantas, and
  G.~C. Nikiforidis, ``Automatic vessel lumen segmentation and stent strut
  detection in intravascular optical coherence tomography,'' \emph{Medical
  physics}, vol.~39, no.~1, pp. 503--513, 2012.

\bibitem{roy2016lumen}
A.~G. Roy, S.~Conjeti, S.~G. Carlier, P.~K. Dutta, A.~Kastrati, A.~F. Laine,
  N.~Navab, A.~Katouzian, and D.~Sheet, ``Lumen segmentation in intravascular
  optical coherence tomography using backscattering tracked and initialized
  random walks,'' \emph{IEEE journal of biomedical and health informatics},
  vol.~20, no.~2, pp. 606--614, 2016.

\bibitem{lu2012automatic}
H.~Lu, M.~Gargesha, Z.~Wang, D.~Chamie, G.~F. Attizzani, T.~Kanaya, S.~Ray,
  M.~A. Costa, A.~M. Rollins, H.~G. Bezerra \emph{et~al.}, ``Automatic stent
  detection in intravascular oct images using bagged decision trees,''
  \emph{Biomed Opt Express}, vol.~3, no.~11, pp. 2809--2824, 2012.

\bibitem{wang20153}
Z.~Wang, M.~W. Jenkins, G.~C. Linderman, H.~G. Bezerra, Y.~Fujino, M.~A. Costa,
  D.~L. Wilson, and A.~M. Rollins, ``3-d stent detection in intravascular oct
  using a bayesian network and graph search,'' \emph{IEEE Trans Med Imaging},
  vol.~34, no.~7, pp. 1549--1561, 2015.

\bibitem{yabushita2002characterization}
H.~Yabushita, B.~E. Bouma, S.~L. Houser, H.~T. Aretz, I.-K. Jang, K.~H.
  Schlendorf, C.~R. Kauffman, M.~Shishkov, D.-H. Kang, E.~F. Halpern
  \emph{et~al.}, ``Characterization of human atherosclerosis by optical
  coherence tomography,'' \emph{Circulation}, vol. 106, no.~13, pp. 1640--1645,
  2002.

\bibitem{wang2012volumetric}
Z.~Wang, D.~Chamie, H.~G. Bezerra, H.~Yamamoto, J.~Kanovsky, D.~L. Wilson,
  M.~A. Costa, and A.~M. Rollins, ``Volumetric quantification of fibrous caps
  using intravascular optical coherence tomography,'' \emph{Biomed Opt
  Express}, vol.~3, no.~6, pp. 1413--1426, 2012.

\bibitem{zahnd2015quantification}
G.~Zahnd, A.~Karanasos, G.~van Soest, E.~Regar, W.~Niessen, F.~Gijsen, and
  T.~van Walsum, ``Quantification of fibrous cap thickness in intracoronary
  optical coherence tomography with a contour segmentation method based on
  dynamic programming,'' \emph{Int J Comput Assist Radiol Surg}, vol.~10,
  no.~9, pp. 1383--1394, 2015.

\bibitem{mehanna2013volumetric}
E.~Mehanna, H.~G. Bezerra, D.~Prabhu, E.~Brandt, D.~Chami{\'e}, H.~Yamamoto,
  G.~F. Attizzani, S.~Tahara, N.~Van~Ditzhuijzen, Y.~Fujino \emph{et~al.},
  ``Volumetric characterization of human coronary calcification by
  frequency-domain optical coherence tomography,'' \emph{Circulation Journal},
  vol.~77, no.~9, pp. 2334--2340, 2013.

\bibitem{mintz2015intravascular}
G.~S. Mintz, ``Intravascular imaging of coronary calcification and its clinical
  implications,'' \emph{JACC: Cardiovascular Imaging}, vol.~8, no.~4, pp.
  461--471, 2015.

\bibitem{xu2008characterization}
C.~Xu, J.~M. Schmitt, S.~G. Carlier, and R.~Virmani, ``Characterization of
  atherosclerosis plaques by measuring both backscattering and attenuation
  coefficients in optical coherence tomography,'' \emph{Journal of biomedical
  optics}, vol.~13, no.~3, p. 034003, 2008.

\bibitem{van2010atherosclerotic}
G.~Van~Soest, T.~P. Goderie, E.~Regar, S.~Koljenovic, A.~G.~J. van Leenders,
  N.~Gonzalo, S.~van Noorden, T.~Okamura, B.~E. Bouma, G.~J. Tearney
  \emph{et~al.}, ``Atherosclerotic tissue characterization in vivo by optical
  coherence tomography attenuation imaging,'' \emph{Journal of biomedical
  optics}, vol.~15, no.~1, p. 011105, 2010.

\bibitem{vermeer2014depth}
K.~Vermeer, J.~Mo, J.~Weda, H.~Lemij, and J.~De~Boer, ``Depth-resolved
  model-based reconstruction of attenuation coefficients in optical coherence
  tomography,'' \emph{Biomed Opt Express}, vol.~5, no.~1, pp. 322--337, 2014.

\bibitem{gargesha2015parameter}
M.~Gargesha, R.~Shalev, D.~Prabhu, K.~Tanaka, A.~M. Rollins, M.~Costa, H.~G.
  Bezerra, and D.~L. Wilson, ``Parameter estimation of atherosclerotic tissue
  optical properties from three-dimensional intravascular optical coherence
  tomography,'' \emph{J Med Imaging}, vol.~2, no.~1, p. 016001, 2015.

\bibitem{liu2017tissue}
S.~Liu, Y.~Sotomi, J.~Eggermont, G.~Nakazawa, S.~Torii, T.~Ijichi, Y.~Onuma,
  P.~W. Serruys, B.~P. Lelieveldt, and J.~Dijkstra, ``Tissue characterization
  with depth-resolved attenuation coefficient and backscatter term in
  intravascular optical coherence tomography images,'' \emph{Journal of
  biomedical optics}, vol.~22, no.~9, p. 096004, 2017.

\bibitem{ughi2013automated}
G.~J. Ughi, T.~Adriaenssens, P.~Sinnaeve, W.~Desmet, and J.~D’hooge,
  ``Automated tissue characterization of in vivo atherosclerotic plaques by
  intravascular optical coherence tomography images,'' \emph{Biomed Opt
  Express}, vol.~4, no.~7, pp. 1014--1030, 2013.

\bibitem{celi2014vivo}
S.~Celi and S.~Berti, ``In-vivo segmentation and quantification of coronary
  lesions by optical coherence tomography images for a lesion type definition
  and stenosis grading,'' \emph{Medical image analysis}, vol.~18, no.~7, pp.
  1157--1168, 2014.

\bibitem{athanasiou2014methodology}
L.~S. Athanasiou, C.~V. Bourantas, G.~Rigas, A.~I. Sakellarios, T.~P. Exarchos,
  P.~K. Siogkas, A.~Ricciardi, K.~K. Naka, M.~I. Papafaklis, L.~K. Michalis
  \emph{et~al.}, ``Methodology for fully automated segmentation and plaque
  characterization in intracoronary optical coherence tomography images,''
  \emph{Journal of biomedical optics}, vol.~19, no.~2, p. 026009, 2014.

\bibitem{abdolmanafi2017deep}
A.~Abdolmanafi, L.~Duong, N.~Dahdah, and F.~Cheriet, ``Deep feature learning
  for automatic tissue classification of coronary artery using optical
  coherence tomography,'' \emph{Biomed Opt Express}, vol.~8, no.~2, pp.
  1203--1220, 2017.

\bibitem{Krizhevsky.2012}
A.~Krizhevsky, I.~Sutskever, and G.~E. Hinton, ``{ImageNet Classification with
  Deep Convolutional Neural Networks},'' in \emph{{Advances in Neural
  Information Processing Systems 25}}, {F. Pereira}, {C. J. C. Burges}, {L.
  Bottou}, and {K. Q. Weinberger}, Eds.\hskip 1em plus 0.5em minus 0.4em\relax
  {Curran Associates, Inc}, 2012, pp. 1097--1105.

\bibitem{Long.2015}
J.~Long, E.~Shelhamer, and T.~Darrell, ``{Fully convolutional networks for
  semantic segmentation},'' in \emph{{Proceedings of the IEEE CVPR}}, 2015, pp.
  3431--3440.

\bibitem{esteva2017dermatologist}
A.~Esteva, B.~Kuprel, R.~A. Novoa, J.~Ko, S.~M. Swetter, H.~M. Blau, and
  S.~Thrun, ``Dermatologist-level classification of skin cancer with deep
  neural networks,'' \emph{Nature}, vol. 542, no. 7639, p. 115, 2017.

\bibitem{pereira2016brain}
S.~Pereira, A.~Pinto, V.~Alves, and C.~A. Silva, ``Brain tumor segmentation
  using convolutional neural networks in mri images,'' \emph{IEEE Trans Med
  Imaging}, vol.~35, no.~5, pp. 1240--1251, 2016.

\bibitem{litjens2017survey}
G.~Litjens, T.~Kooi, B.~E. Bejnordi, A.~A.~A. Setio, F.~Ciompi, M.~Ghafoorian,
  J.~A. van~der Laak, B.~van Ginneken, and C.~I. S{\'a}nchez, ``A survey on
  deep learning in medical image analysis,'' \emph{Medical image analysis},
  vol.~42, pp. 60--88, 2017.

\bibitem{he2018convolutional}
S.~He, J.~Zheng, A.~Maehara, G.~Mintz, D.~Tang, M.~Anastasio, and H.~Li,
  ``Convolutional neural network based automatic plaque characterization for
  intracoronary optical coherence tomography images,'' in \emph{Medical Imaging
  2018: Image Processing}, vol. 10574, 2018, p. 1057432.

\bibitem{badrinarayanan2017segnet}
V.~Badrinarayanan, A.~Kendall, and R.~Cipolla, ``Segnet: A deep convolutional
  encoder-decoder architecture for image segmentation,'' \emph{IEEE
  transactions on pattern analysis and machine intelligence}, vol.~39, no.~12,
  pp. 2481--2495, 2017.

\bibitem{oliveira2018coronary}
D.~A. Oliveira, M.~M. Macedo, P.~Nicz, C.~Campos, P.~Lemos, and M.~A.
  Gutierrez, ``Coronary calcification identification in optical coherence
  tomography using convolutional neural networks,'' in \emph{Medical Imaging
  2018: Biomedical Applications in Molecular, Structural, and Functional
  Imaging}, vol. 10578, 2018, p. 105781Y.

\bibitem{boi2018survey}
A.~Boi, A.~D. Jamthikar, L.~Saba, D.~Gupta, A.~Sharma, B.~Loi, J.~R. Laird,
  N.~N. Khanna, and J.~S. Suri, ``A survey on coronary atherosclerotic plaque
  tissue characterization in intravascular optical coherence tomography,''
  \emph{Current atherosclerosis reports}, vol.~20, no.~7, p.~33, 2018.

\bibitem{rico2016automatic}
J.~J. Rico-Jimenez, D.~U. Campos-Delgado, M.~Villiger, K.~Otsuka, B.~E. Bouma,
  and J.~A. Jo, ``Automatic classification of atherosclerotic plaques imaged
  with intravascular oct,'' \emph{Biomed Opt Express}, vol.~7, no.~10, pp.
  4069--4085, 2016.

\bibitem{He.2016}
K.~He, X.~Zhang, S.~Ren, and J.~Sun, ``{Deep residual learning for image
  recognition},'' in \emph{{Proceedings of the IEEE CVPR}}, 2016, pp. 770--778.

\bibitem{Szegedy.2016b}
C.~Szegedy, V.~Vanhoucke, S.~Ioffe, J.~Shlens, and Z.~Wojna, ``{Rethinking the
  inception architecture for computer vision},'' in \emph{{Proceedings of the
  IEEE CVPR}}, 2016, pp. 2818--2826.

\bibitem{shin2016deep}
H.-C. Shin, H.~R. Roth, M.~Gao, L.~Lu, Z.~Xu, I.~Nogues, J.~Yao, D.~Mollura,
  and R.~M. Summers, ``Deep convolutional neural networks for computer-aided
  detection: Cnn architectures, dataset characteristics and transfer
  learning,'' \emph{IEEE Trans Med Imaging}, vol.~35, no.~5, pp. 1285--1298,
  2016.

\bibitem{gessert2018}
N.~Gessert, M.~Heyder, S.~Latus, M.~Lutz, and A.~Schlaefer, ``Plaque
  classification in coronary arteries from ivoct images using convolutional
  neural networks and transfer learning,'' in \emph{Computer Assisted Radiology
  and Surgery Proceedings of the 32nd International Congress and
  Exhibition}.\hskip 1em plus 0.5em minus 0.4em\relax Springer, 2018.

\bibitem{tajbakhsh2016convolutional}
N.~Tajbakhsh, J.~Y. Shin, S.~R. Gurudu, R.~T. Hurst, C.~B. Kendall, M.~B.
  Gotway, and J.~Liang, ``Convolutional neural networks for medical image
  analysis: Full training or fine tuning?'' \emph{IEEE Trans Med Imaging},
  vol.~35, no.~5, pp. 1299--1312, 2016.

\bibitem{huang2017densely}
G.~Huang, Z.~Liu, K.~Q. Weinberger, and L.~van~der Maaten, ``Densely connected
  convolutional networks,'' in \emph{Proceedings of the IEEE CVPR}, vol.~1,
  no.~2, 2017, p.~3.

\bibitem{He.2016b}
K.~He, X.~Zhang, S.~Ren, and J.~Sun, ``{Identity mappings in deep residual
  networks},'' in \emph{{ECCV}}, 2016, pp. 630--645.

\bibitem{Lin.2013}
M.~Lin, Q.~Chen, and S.~Yan, ``{Network in network},'' in \emph{{International
  Conference on Learning Representations}}, 2014.

\bibitem{Kamnitsas.2017}
K.~Kamnitsas, C.~Ledig, V.~F.~J. Newcombe, J.~P. Simpson, A.~D. Kane, D.~K.
  Menon, D.~Rueckert, and B.~Glocker, ``{Efficient multi-scale 3D CNN with
  fully connected CRF for accurate brain lesion segmentation},'' \emph{{Medical
  Image Analysis}}, vol.~36, pp. 61--78, 2017.

\bibitem{gessert2018force}
N.~Gessert, J.~Beringhoff, C.~Otte, and A.~Schlaefer, ``{Force estimation from
  OCT volumes using 3D CNNs},'' \emph{{International Journal of Computer
  Assisted Radiology and Surgery}}, vol.~13, no.~7, pp. 1073--–1082, 2018.

\bibitem{Kingma.2014}
D.~Kingma and J.~Ba, ``{Adam: A method for stochastic optimization},'' in
  \emph{{ICLR}}, 2014.

\bibitem{Abadi.2016}
M.~Abadi, A.~Agarwal, P.~Barham, E.~Brevdo, Z.~Chen, C.~Citro, G.~S. Corrado,
  A.~Davis, J.~Dean, and M.~Devin, ``{Tensorflow: Large-scale machine learning
  on heterogeneous distributed systems},'' \emph{{arXiv preprint
  arXiv:1603.04467}}, 2016.

\bibitem{Yosinski.2014}
J.~Yosinski, J.~Clune, Y.~Bengio, and H.~Lipson, ``{How transferable are
  features in deep neural networks?}'' in \emph{{Advances in Neural Information
  Processing Systems}}, 2014, pp. 3320--3328.

\end{thebibliography}
%








\end{document}